\documentclass{article} 
\usepackage{iclr2024_conference,times}

\usepackage{amsmath,amsfonts,bm}

\def\eqref#1{equation~\ref{#1}}

\def\1{\bm{1}}

\DeclareMathAlphabet{\mathsfit}{\encodingdefault}{\sfdefault}{m}{sl}
\SetMathAlphabet{\mathsfit}{bold}{\encodingdefault}{\sfdefault}{bx}{n}

\DeclareMathOperator*{\argmin}{arg\,min}

\usepackage{hyperref}
\usepackage{url}
\usepackage{graphicx}
\usepackage{kotex}
\usepackage{booktabs}
\usepackage{comment}

\newcommand{\nj}[1]{\textcolor{red}{#1}}

\newcommand{\yr}[1]{\textcolor{magenta}{#1}}

\title{Decompose the model: mechanistic interpretability in image models with generalized integrated gradients (GIG)}

\author{Yearim Kim$^{*}$, \quad Sangyu Han\thanks{Equal contribution. $^{\dagger}$Corresponding author. }, \quad Sangbum Han, \quad Nojun Kwak$^{\dagger}$ \\
Department of Computer Science\\
Seoul National University\\
Seoul, 08826, Korea 
\\
\texttt{\{yerim1656, acoexist96, snuhsb,nojunk\}@snu.ac.kr}
}

\iclrfinalcopy 
\begin{document}

\maketitle


\begin{abstract}
In the field of eXplainable AI (XAI) in language models, the progression from local explanations of individual decisions to global explanations with high-level concepts has laid the groundwork for mechanistic interpretability, which aims to decode the exact operations. 
However, this paradigm has not been adequately explored in image models, where existing methods have primarily focused on class-specific interpretations.
This paper introduces a novel approach to systematically trace the entire pathway from input through all intermediate layers to the final output within the whole dataset.
We utilize Pointwise Feature Vectors (PFVs) and Effective Receptive Fields (ERFs) to decompose model embeddings into interpretable Concept Vectors.
Then, we calculate the relevance between concept vectors with our Generalized Integrated Gradients (GIG), enabling a comprehensive, dataset-wide analysis of model behavior.
We validate our method of concept extraction and concept attribution in both qualitative and quantitative evaluations.
Our approach advances the understanding of semantic significance within image models, offering a holistic view of their operational mechanics.
\end{abstract}

\section{Introduction}

In the field of eXplainable AI (XAI), efforts have historically transitioned from Local explanation to Global explanation to Mechanistic Interpretability. 
While local explanation methods including \citet{gradcam, Montavon2017lrp, sundararajan2017ig, han2024respect} have focused on explaining specific decisions for individual instances, global explanation methods seek to uncover overall patterns and behaviors applicable across the entire dataset
\citep{wu2022cpm,xuanyuan2023gnn,singh2024rethinking}.
One step further, mechanistic interpretability methods seek to analyze the fundamental components of the models and provide a holistic explanation of operational mechanics across various layers.

Recently, researchers in language models, especially in Large Language Models (LLMs), have extensively studied mechanistic interpretability to reveal the causal relationships and precise mechanisms transforming inputs into outputs \citep{geva2022TransformerFFL, bricken2023monosemanticity, gurnee2024gpt2}.
Researches on interpretability in image models, however, have typically focused on class-wise explanations \citep{fel2023craft, ghorbani2019ace} rather than dataset-wide explanations. 
Furthermore, mechanistic interpretability, a comprehensive explanation dealing with every layer within the whole dataset, has not been applied to image models.
This distinction arises because images consist of pixels that do not inherently represent concepts, unlike languages where each comprising word itself can be treated as a concept.
Additionally, as meaningful structures in images are localized and only occupy small regions of the entire image, the embedding space in image datasets is far more sparse compared to that in language datasets.

\begin{figure}[ht!]
    \centering
    \includegraphics[width=\linewidth]{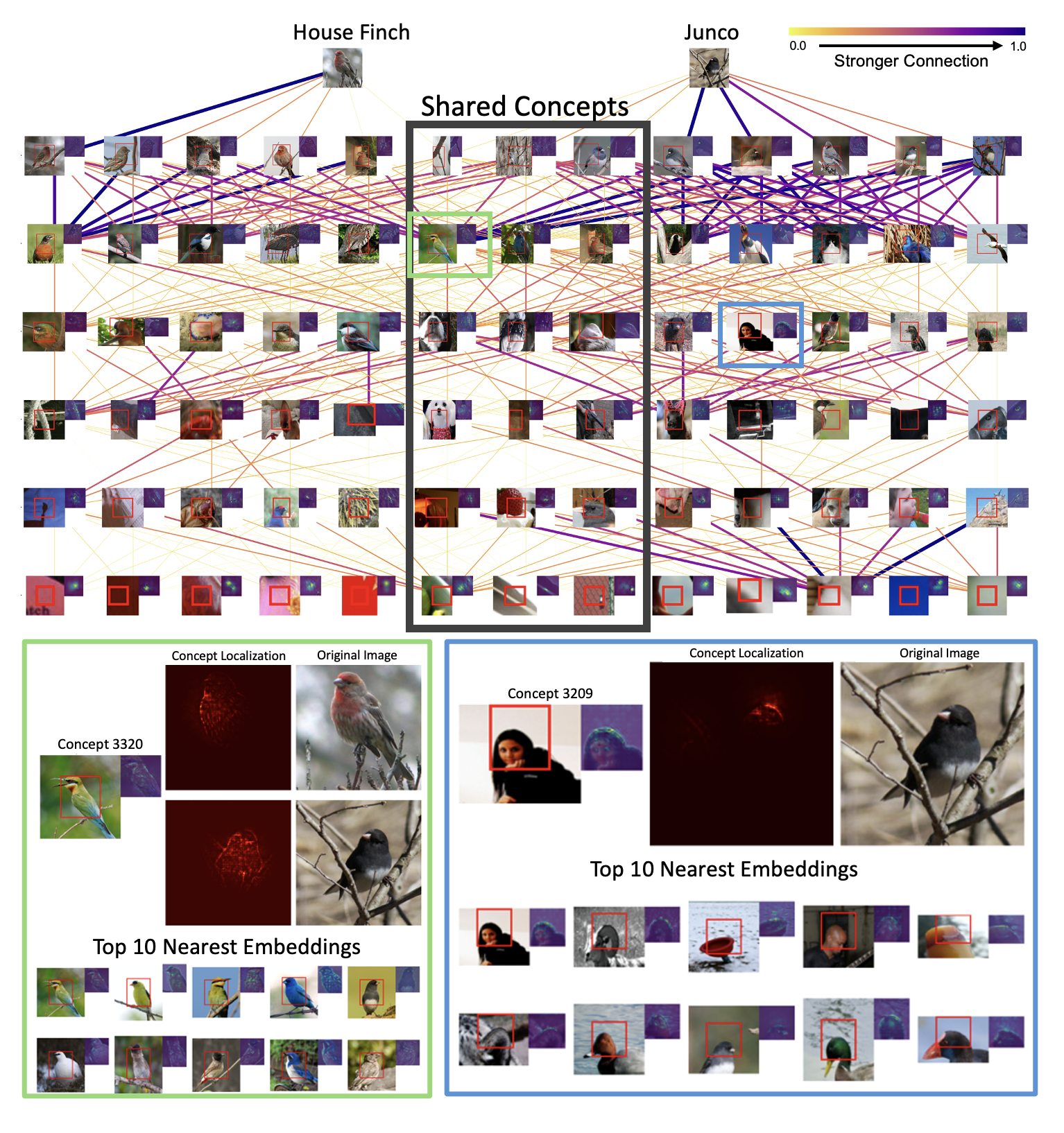}
    \caption{\textbf{Top}: Causal explanation graph from high to low layers. From top to bottom, [Classifier, Layer4.2, Layer3.5, Layer3.2, Layer3.0, Layer2.3, Layer1.2], the bottleneck blocks in ResNet50. The thicker and bluer the edge, the stronger the contribution between concepts. 
    Unlike class-wise global explanation, our method can explain the `Shared concepts' between similar classes. Among the thousands of concepts in a layer, the graph only shows the top-5 most important concepts and top-3 shared concepts.
    \textbf{Bottom left}: Detailed concept visualization of Concept 3,320 
    ``Bird chest" at Layer3.5 Block. With the top 10 nearest embeddings, we can observe Concept 3,320 is ``Bird chest." 
    With its concept localization image, we can effectively see where concept 3,320 resides in input images of the classes, house finch and junco, respectively. \textbf{Bottom right}: Concept 3,209 ``Round head" at Layer3.2 Block. The top-1 representation image of concept 3,209 (A girl's head) seems irrelevant to the class, junco bird. However, with the concept localization image and the corresponding top-10 nearest embeddings, we can see that Concept 3,209 represents the round head of various objects.}
    \label{fig:main}
\end{figure}

In this paper, we present a novel approach to mechanistic interpretation in image models by systematically decomposing and tracing the pathways from input to output across an entire dataset.
Unlike previous methods that often focus on individual classes or specific features, our approach provides a comprehensive, dataset-wide understanding of the entire model's behavior.
This is the first to explain the model's embedding within the whole dataset, throughout the whole layers (See Figure.~\ref{fig:main}).
We decompose the model's embedding with the dataset-wide concept vectors, enabling the existence of ``Shared Concepts'' unlike other class-wise C-XAI methods \citep{fel2023craft}, \citep{kowal2024vcc}.
We also introduce \textbf{Generalized Integrated Gradients} to enable explaining the causal relationships between the concepts of different layers.

Following the framework of \cite{han2024respect}, we use the Pointwise Feature Vector (PFV) for our analysis unit.
A PFV is defined as the channel-axis pre-activation vector of each layer, serving as the fundamental unit encoding the network's representations.
We further utilize the instance-level Effective Receptive Fields (ERFs) of the PFVs to label their semantic meaning, enabling a direct investigation of the PFV vector space. 
By leveraging ERF, we aim to identify clear and meaningful linear bases within the PFV vector space, allowing us to decompose previously unknown PFVs into interpretable principal components, which we refer to as Concept Vectors (CVs).

To find out the bases and identify the CVs, we leverage several clustering methods including dictionary learning, k-means, and Sparse AutoEncoder.
Among the clustering methods, we employ bisecting k-means clustering, since it is well-suited for the PFV vector space, which is highly sparse and variably dense.
For instance, background features often cluster densely, while critical features, such as ``the beak of a bird", may occupy a broader, less dense area.

To find out the causal relationships between concept vectors in different layers, we introduce \textbf{Generalized Integrated Gradient} (GIG), which effectively captures interlayer contributions.
By combining Concept Vectors with GIG, we can offer a comprehensive causal analysis of the ResNet50 model, from the lowest layers to the final class predictions.
Fig.~\ref{fig:main} shows examples of how our method iteratively aggregates concepts in the previous layer to form a higher-layer abstract concept.

In this study, we focus on the ResNet50 architecture.
Yet, our approach is not limited to convolutional architectures and also universally applicable across various modalities, including transformer architectures, as convolutional layers can be viewed as a special case of multi-head attention with fixed attention \citep{cordonnier2020relationship}. 
Our framework facilitates a deeper understanding of the semantic significance of features, thus advancing the mechanistic interpretability image models.

\section{Related Works}

\textbf{Local Explanation} methods, such as LIME \citep{ribeiro2016should} and SHAP \citep{lundberg2017unified}, have been widely studied to provide an explanation for a specific instance.
These techniques identify the most influential input features for specific instances, aiding in the understanding of particular decisions.
However, they often lack generalizability, as they focus on isolated cases rather than providing insights into the model's overall behavior across different inputs.

To address these limitations, \textbf{Global Explanation} methods were introduced \citep{kim2018tcav, ghorbani2019ace}.
Unlike local explanations, they aim to provide a broad understanding of a model's behavior by analyzing patterns and decision-making processes across the entire dataset.
However, in vision models, global explanation methods, such as TCAV 
\citep{kim2018tcav} and ACE 
\citep{ghorbani2019ace}, remain class-specific.
These methods quantify the importance of high-level concepts or identify concepts through multi-resolution segmentation and clustering techniques, but they often focus on understanding these concepts within individual classes.
This class-specific focus limits their ability to provide a comprehensive understanding of the model’s behavior across the entire dataset, which is a crucial component of true global explanation.
Moreover, \cite{fel2023holistic} extends various local, attribution methods into global, Concept ATtribution (CAT) methods, but it primarily focuses on interpreting the output from a single layer without addressing the interpretation of interactions between different layers.
This limitation underscores the importance of our approach, which provides a more comprehensive analysis by considering the relationships between layers.

As the need for deeper insights into model behavior grew, researchers turned to \textbf{Mechanistic Interpretability}.
While global explanations provide a high-level understanding of a model’s behavior by identifying patterns across a dataset, they do not analyze the internal workings of the model.
Mechanistic Interpretability addresses this gap by focusing on how specific components—such as neurons, layers, or circuits—interact and contribute to the overall function.
For instance, CRP \citep{Achtibat2023CRP}, an extension of LRP \citep{Montavon2017lrp}, provides detailed exploration of how the concepts impact the model's output at each layer, by introducing concept-conditional relevance mapping. 
VCC \citep{kowal2024vcc}, an extension of TCAV \citep{kim2018tcav}, interprets how individual concepts contribute to the model's decisions across different layers. 
However, both of them focus on a specific class, thereby limiting their capacities to offer a more comprehensive and generalized view of the model's behavior across the entire dataset.
In contrast, our approach broadens the inter-layer analysis to include the entire dataset, enabling a more thorough examination of the model's decision-making process.
Our work is the first application of this technique to vision models, uncovering complex interactions that traditional global methods and class-specific approaches may overlook.

\section{Method}
\begin{figure*}[t]
    \centering
    \includegraphics[width=\linewidth]{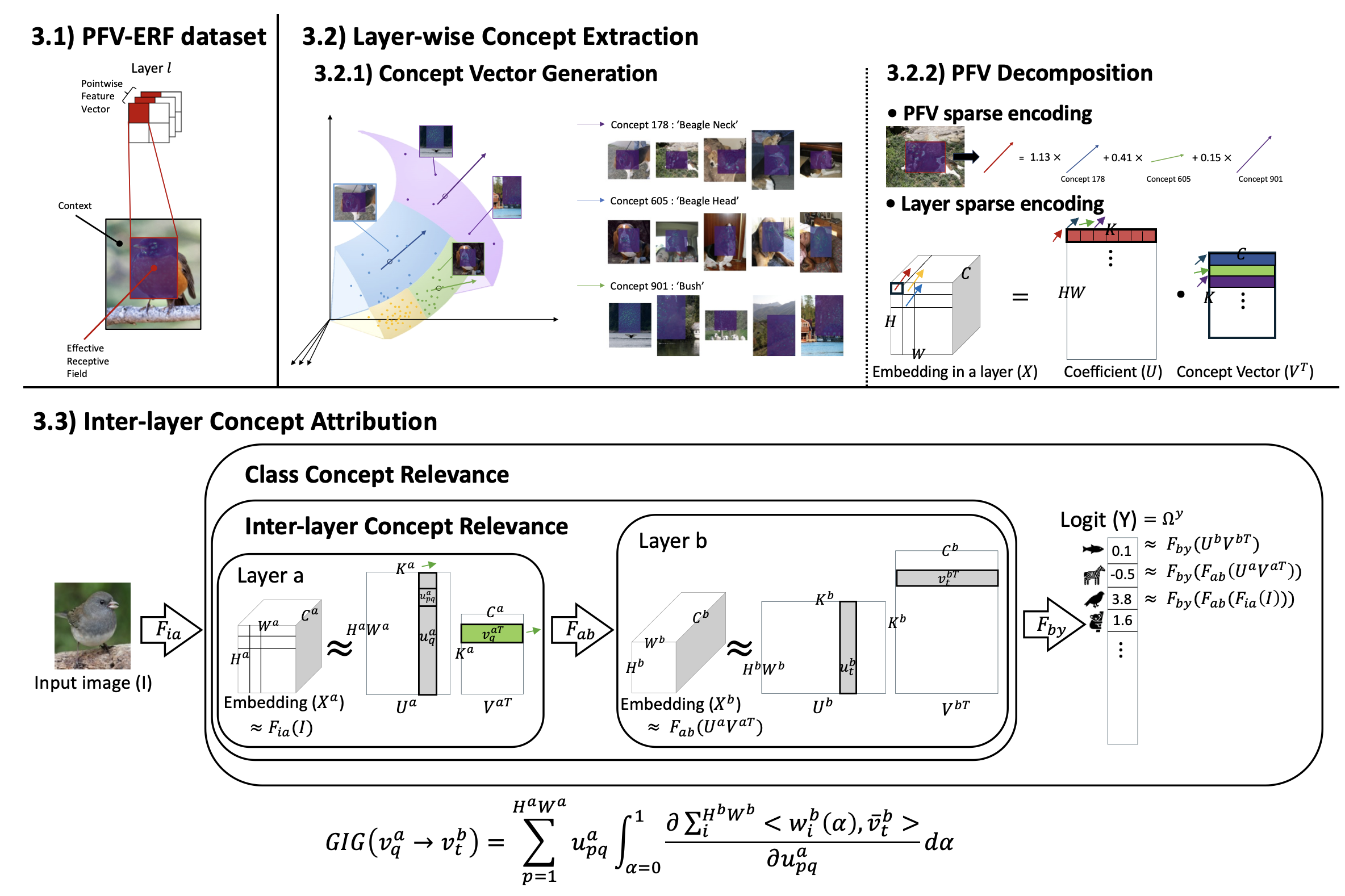}
    \caption{Method overview. \textbf{3.1)} Our dataset. The Pointwise Feature Vector (PFV) in the hidden layer is assigned a meaning by labeling it with the Effective Receptive Field (ERF). The ERF image (the blue area in the picture) uses a single color to represent the importance, making it difficult to interpret. Therefore, the context portion, around the ERF, is added to make the significance of the ERF more understandable. \textbf{3.2)} Layer-wise concept extraction. \textbf{3.2.1)} The PFV vector space exhibits a diverse density, with high density around specific concepts and sparsity elsewhere. Hence, bisecting clustering, suitable for such data structures, is employed to extract concept vectors. The meaning of each concept vector is then explained through the sample with the highest cosine similarity to the concept vector. \textbf{3.2.2)} Reconstruct the PFV and embeddings in a layer with the extracted concept vectors. \textbf{3.3)} Inter-layer concept attribution, employing Generalized Integrated Gradients (GIG). 
    }
    \label{fig:method_overview}
\end{figure*}

Fig.~\ref{fig:method_overview} shows the overview of method. 
In the figure, each process of our method is represented with the corresponding section number. 
\subsection{Analysis Unit: PFV-ERF dataset}
In our study on mechanistic interpretability in image models, we utilize several key components essential for understanding and analyzing the network's behavior from the work of \cite{han2024respect}.
Specifically, we use a Pointwise Feature Vector (PFV) as the unit of analysis and its corresponding Effective Receptive Field (ERF) as the visual label to effectively show and validate the knowledge encoded by the PFV.

Firstly, a PFV is a vector of neurons along the channel axis within a hidden layer that share an identical receptive field. 
Given the embedding of layer $l$ denoted as $A^l \in \mathbb{R}^{H^l W^l\times C^l}$, where $C^l$ is the number of channels and $H^l W^l$ represents the spatial dimensions of the feature map, 
the PFV at position $p \in \{1,\cdots, H^l W^l\}$ is represented as $\mathbf{x}^{l}_p \in \mathbb{R}^{C^l}$.
This vector encapsulates a localized feature representation at a specific point within the input image, providing a clear characterization of the features at that particular location. 
Unlike individual neurons, a PFV ensures monosemanticity, capturing a singular, coherent concept from the multi-channel features at a specific spatial location.
Therefore, we decompose a layer in a network using PFVs.
More specifically, a PFV in the preactivation space is linearly decomposed with the concept vectors. 

Secondly, we use the ERF as the PFV's label.
Receptive Field (RF) denotes the region within the input image that influences the activation of a specific feature, defining the spatial extent over which the input pixels contribute to the feature's activation.
\citet{han2024respect} further refined this concept to Effective Receptive Field (ERF) to highlight the differential impact of individual pixels, identifying those that are most influential in the computation of the PFV.
With ERF, we directly attribute a meaning (or a concept) to each hidden layer feature vector (PFV in our case), in contrast to other existing methods, which infer feature vector meaning through indirect techniques.
\citet{ghorbani2019ace} and \citet{kowal2024vcc} used global average pooling after masking, and \citet{fel2023craft} used bilinear interpolation on the masked feature maps to create a squared region to provide an indirect explanation of feature vectors by transforming segmented areas into representative vectors.
Yet, with ERF, we explicitly assign meanings to the hidden layer feature vectors, treating them as representations of specific concepts so that we can offer a more straightforward interpretation of how particular features contribute to the model's decisions.
\subsection{Concept Extraction}


\subsubsection{Concept vector generation}
To determine the principal axis of the PFVs in each layer and find out the concept vectors, we utilize ImageNet validation dataset, consisting of 50,000 images.
Even though there are $HW$ PFVs within a single layer, we take only one PFV and its corresponding ERF, resulting in 50,000 PFV-ERF pairs per layer for the dataset.
In an image, we sample a PFV in a non-uniform sense to reflect its contribution to the output (logit), due to the foreground-background imbalance problem in images;
If we sample PFVs randomly from an image, then the majority would capture the background, which would be irrelevant to the output class.
For example, the sky in an image could be present across various classes, leading to an overrepresentation of the class-irrelevant feature, sky, rather than critical features like a bird's beak.
This overrepresentation of irrelevant features within the PFVs could skew the identification of the principal axes of the PFVs.
Thus, we probabilistically select a single PFV from each image in proportion to its contribution to the output, ensuring a more balanced dataset and solving the imbalance problem. 

With this balanced PFV-ERF dataset, we employ a bisecting k-means clustering \citep{Steinbach2000ACO}, which iteratively splits the data into two clusters until a predefined number of clusters is reached.
This approach effectively navigates the complex manifold of image data, where some regions are sparse, containing rare or atypical features, while others are dense, filled with frequently encountered features.
After clustering, we assign the centroid of each cluster as a concept vector.

\subsubsection{PFV decomposition}
\label{subsubsec:decomposition}
Let there be $k$ concept vectors in layer $l$, denoted as $\mathbf{v}_1^l, \cdots, \mathbf{v}_k^l$, discovered in the same $C$-dimensional vector space \( \mathcal{V}^l \) with PFVs.
Then, each PFV $\mathbf{x}_p^l$ can be expressed as a linear combination of the concept vectors:
\begin{equation}
    \mathbf{x}^{l}_p = \sum_{j=1}^{k} u_{pj} \mathbf{v}_j^l + \epsilon,
\end{equation}
where $u_{pj}$ is the coefficient representing the contribution of the $j$-th concept vector to PFV $x^{l}_p$ ($\mathbf{u}_p = [u_{pi},\cdots, u_{pk}]^T$), and $\epsilon$ is the residual error.
To determine the coefficients $\mathbf{u}_p$, we use Lasso regression, which minimizes the following objective function:
\begin{equation}
    \mathbf{u}_p^* = \argmin_{\mathbf{u}_p} \left\{ \frac{1}{2} \left\| \mathbf{x}^{l}_p - \sum_{j=1}^{k} u_{pj} \mathbf{v}_j^l \right\|_2^2 + \lambda \sum_{j=1}^{k} |u_{pj}| \right\},
\end{equation}
where $\lambda$ is a regularization parameter that controls the sparsity of the solution, encouraging many of the coefficients $u_{pj}$ to be zero.
By using lasso regression, we can reconstruct the original PFVs with a small number of concept vectors.

In this way, the embeddings in the $l$-th layer, $X^l \in \mathbb{R}^{HW \times C}$, can be approximated by $k$ concept vectors as $\tilde{X}^l = U V^T$ where $U \in \mathbb{R}^{HW \times k}$ is the coefficient matrix and each column of $V \in \mathbb{R}^{C \times k}$ contains a concept vector.

\subsection{Inter-layer concept attribution}
\label{subsec: concept attribution}
In this paper, we leveraged 
Integrated Gradients (IG) \citep{sundararajan2017ig} 
to calculate the inter-layer concept attribution.
Among other attribution methods, we utilized IG due to its superiority across various reliability metrics, such as C-Deletion, C-Insertion, and C-µFidelity, which are crucial in ensuring the robustness and accuracy of concept-based explanations \citep{fel2023holistic}.

Based on IG, we propose a novel method, \textbf{Generalized Integrated Gradients (GIG)}, which extends the integrated gradients to quantify the contribution of a specific concept vector in a layer to both the final class output and the concept vectors of subsequent layers.

Let $a$ and $b$ denote the preceding and target layer, and $X^l  (l \in \{a,b \})$ be the embeddings of the corresponding layer. In this work, we want to measure the influence of a query concept vector in layer $a$, $\mathbf{v}_q^a$, on the target concept vector in layer $b$, $\mathbf{v}_t^b$.
To compute the attribution for the target concept vector, we first compute the output embeddings
\begin{equation}
    \Omega^b (\alpha) = F_{ab}(\alpha U^a V^{aT})
    \label{eq:omega}
\end{equation} 
in layer $b$ by varying the embeddings in layer $a$ from $0$ to $\tilde{X}^a = U^a V^{aT}$, i.e, $\alpha \in [0,1]$ in Eq.~(\ref{eq:omega}).
Here, $F_{ab}$ represents the nonlinear function from layer $a$ to $b$ and $U^a V^{aT}$ is the approximation of ${X}^a$ obtained in Sec. \ref{subsubsec:decomposition}.
Then, we project $\Omega^b(\alpha)$ onto the target concept vector $\mathbf{v}_t^b$ and obtain the projected vectors. These projected vectors are spatially aggregated and we compute the integrated gradient for the $q$-th element of the coefficient vector, $u_{pq}$, which is the component of $\mathbf{v}^a_q$ in the PFV $\mathbf{x}_p^a$ at position $p$ as follows:
\begin{equation}
    \text{GIG}(\mathbf{v}_q^a|_p  \rightarrow \mathbf{v}^b_t) = u_{pq}^a \int_{\alpha=0} ^ {1}\frac{\partial \sum_{i=1}^{H^b W^b} \langle \mathbf{w}_i^b(\alpha), \bar{\mathbf{v}}_{t}^b \rangle}{\partial u_{pq}^a}  d \alpha.
\end{equation}
Here, $\bar{\mathbf{v}}_{t}^b$ is the normalized version of $\mathbf{v}_{t}^b$, $\langle \cdot, \cdot \rangle$ is the inner product operation and $\mathbf{w}_i^b$ is the embedding of $\Omega^b$ at the $i$-th position.  

Note that the above GIG measures the attribution of the query concept vector at position $p$ to the target concept vector in a subsequent layer. 
To measure the attribution of a query concept vector, $\mathbf{v}_q^a$, to the target concept vector, $\mathbf{v}^b_t$, we sum up all the attributions of $\mathbf{v}_q^a$ at different positions as follows:
\begin{equation}
    \text{GIG}(\mathbf{v}^a_q \rightarrow \mathbf{v}^b_t) := \sum_{p=1}^{H^a W^a} \text{GIG}(\mathbf{v}_q^a|_p \rightarrow \mathbf{v}^b_t) = \sum_{p=1}^{H^a W^a} u_{pq}^a \int_{\alpha=0} ^ {1}\frac{\partial \sum_{i=1}^{H^b W^b} \langle \mathbf{w}_i^b(\alpha), \bar{\mathbf{v}}_{t}^b \rangle}{\partial u_{pq}^a}  d \alpha.
    \label{eq:GIG}
\end{equation}

\textbf{Class Concept Relevance \ }
To quantify the class importance score of a query concept vector in layer $a$, $\mathbf{v}_q^a$, for the final output (contribution of the concept vector to the given class), we treat each class label $c$ as an independent concept.
Thus, we convert the class index into a one-hot vector, $\mathbf{e}_c \in \{0,1\}^{N}$, where $N$ is the number of classes: 
\begin{equation}
(\mathbf{e}_c)_i = \begin{cases} 
1 & \text{if } i = c \\
0 & \text{if } i \neq c.
\end{cases}
\end{equation}

Then, we calculate the class contribution of the concept vector, $\mathbf{v}^a_q$ using Eq.~(\ref{eq:GIG}) as follows: 
\begin{equation}
    \text{GIG}(\mathbf{v}^a_q \rightarrow \mathbf{e}_c) := \sum_{p=1}^{H^a W^a} \text{GIG}(\mathbf{v}_q^a|_p \rightarrow \mathbf{e}_c) = \sum_{p=1}^{H^a W^a} u_{pq}^a \int_{\alpha=0} ^ {1}\frac{\partial \langle \mathbf{w}^y(\alpha), \mathbf{e}_{c} \rangle}{\partial u_{pq}^a}  d \alpha,
    \label{eq:GIG_class}
\end{equation}
where $y$ indicates the output layer and $\mathbf{w}^y(\alpha)$ is the predicted class probability vector for input embeddings scaled by $\alpha$, i.e, $\mathbf{w}^y=F_{ay}(\alpha U^a V^{aT})$.

By employing our Generalized Integrated Gradients, we aim to uncover the mechanistic interpretability in image models, providing a detailed understanding of how these networks process image data and construct specific concepts through the layers.
\section{Experiment}
To demonstrate the effectiveness of our method, we provide two kinds of qualitative analysis including one-class explanation (Fig.~\ref{fig:qual_img}) and two-class explanation (Fig.~\ref{fig:main}).
Furthermore, we validate our method of concept extraction and concept attribution with comprehensive experiments in Sec.~\ref{val}.

\textbf{Settings.}
Following \citet{bricken2023monosemanticity}, we selected the concept size of each layer as 8 times the number of channels in that layer, making overcomplete linear basis. 
For classic dictionary learning, we utilized the Least Angle Regression (LARS) algorithm and the Lasso LARS algorithm for PFV decomposition.
For sparse autoencoder, we followed the setting of 
\citet{templeton2024scaling}.
For both methods, we extend them by decomposing PFVs directly into coefficients and concept vectors without relying on global average pooling, as they have been applied either at the token level within Transformer architecture, or on the global average pooled outputs of ResNet50 architecture.

\subsection{Qualitative Analysis}
\begin{figure*}[t]
    \centering
    \includegraphics[width=\linewidth]{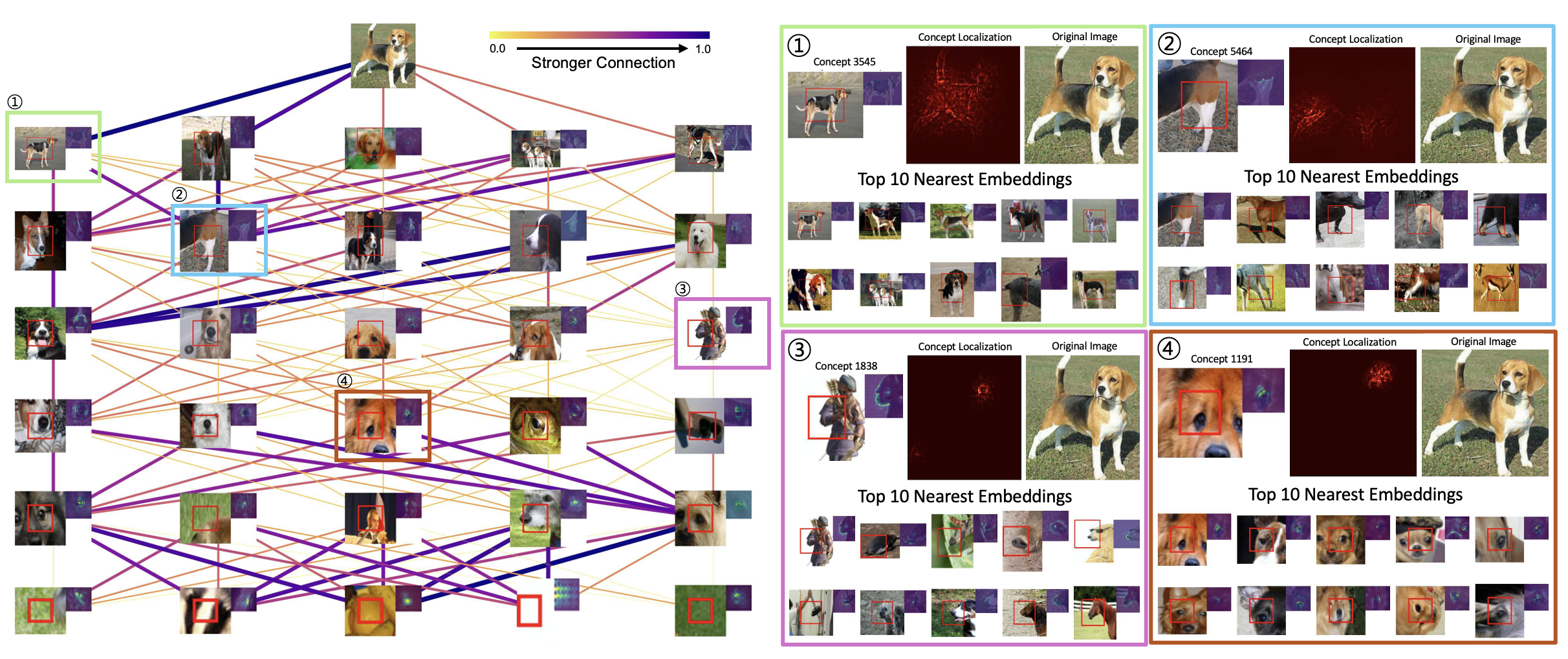}
    \caption{\textbf{Left}: Causal explanation graph of `Foxhound'. From top to bottom, [Classifier, Layer4.2, Layer3.5, Layer3.2, Layer3.0, Layer2.3, Layer1.2], the bottleneck blocks in ResNet50. Our method can also provide a dataset-wide explanation of a single image. \textbf{Right}: Detailed concept visualization of the colored boxes, 1) green, 2) blue, 3) purple, and 4) brown. \textcircled{1} Concept 3,545 ``Dog Body" at Layer4.2 Block. \textcircled{2} Concept 5,464 ``Dog Leg" at Layer3.5 Block. \textcircled{3} Concept 1,838 ``Rounded Cone" at Layer3.2 Block. With Top-1 representation image of `folded arm', the concept seems irrelevant to the input image. However, the concept localization and the top-10 nearest embeddings show that Concept 1,838 represents ``Rounded Cone". \textcircled{4} Concept 1,191 ``Eye" at Layer2.3 Block. Best viewed when enlarged.}
    \label{fig:qual_img}
\end{figure*}

As seen in Fig.~\ref{fig:qual_img}, we can explain how the concept components are constructed through layers.
Moreover, as shown in Fig.~\ref{fig:main}, we can even find out the shared concepts, since we analyze the models within the whole dataset, not a specific class.

\subsection{Validation of our method}
\label{val}
Since our method involves two main steps, we validate the steps of our method with both qualitative and quantitative experiments: Sec.~\ref{val: extraction} for Concept Extraction, and Sec.~\ref{val: attribution} for Inter-layer Concept Attribution.

\subsubsection{Validation of Concept Extraction}
\label{val: extraction}
\begin{figure*}[t]
    \centering
    \includegraphics[width=\linewidth]{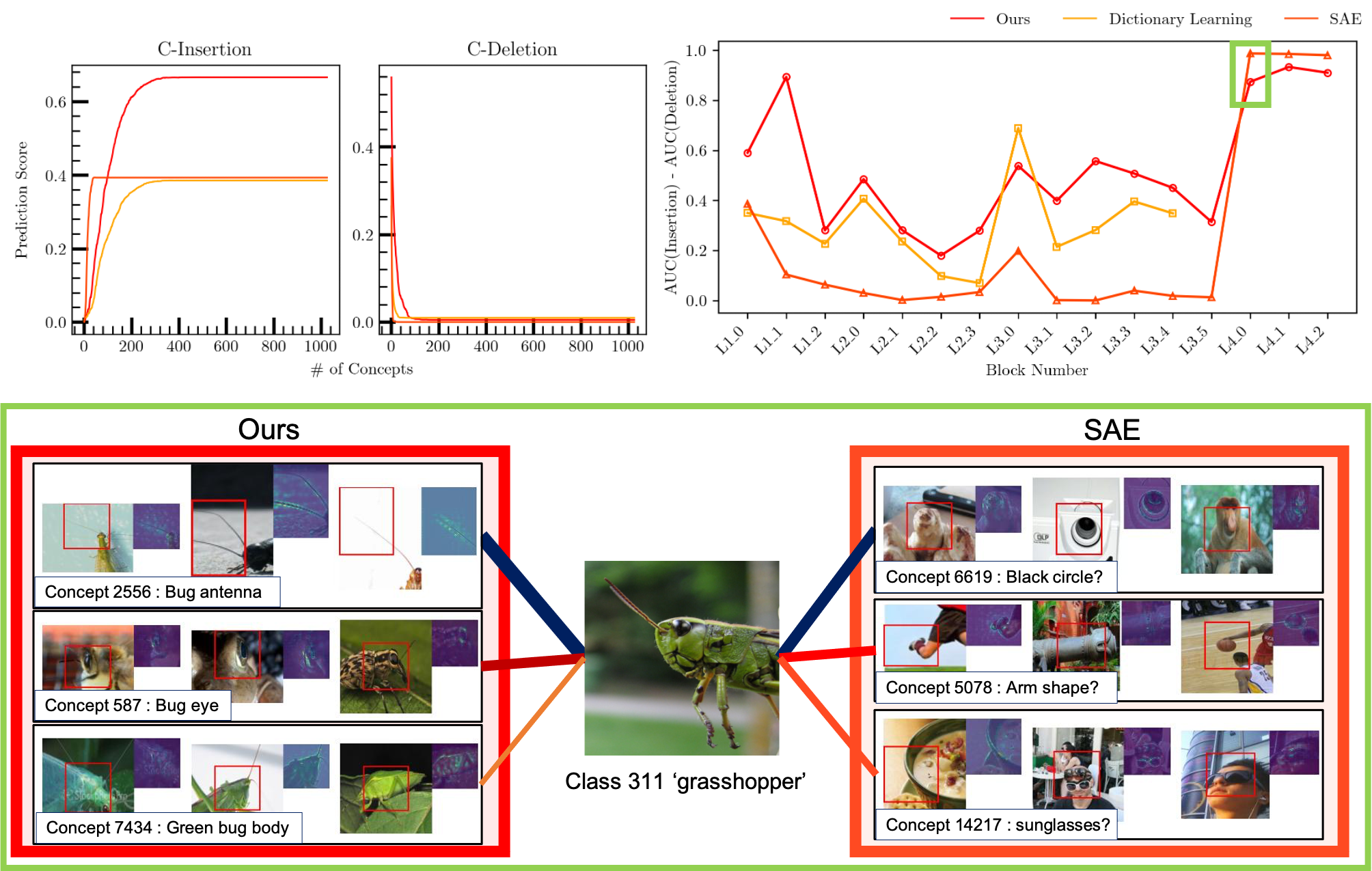}
    \caption{Validation of Concept Extraction. \textbf{Top Left}: Comparison of C-Insertion and C-Deletion curves for three concept extraction methods applied to ResNet50's first block of first stage (Layer1.0). In C-Insertion, `Ours' 
    achieves the highest final score, leading to highest AUC. In C-Deletion, `Ours' degrades slowly, due to the highest initial prediction score. \textbf{Top Right}: AUC differences across different block numbers for a balanced comparison, as there is a tendency that the better the insertion performance, the worse the deletion performance. 
    It shows that `Ours' generally outperforms the other methods across various blocks. 
    \textbf{Bottom}: Top 3 most important concepts found by Sparse AutoEncoder (SAE) and `Ours' for classifying grasshopper image at Layer4.0. Even though SAE excels our method in AUC difference on later layers, the concepts extracted by SAE seem less persuasive than those from `Ours'.}
    
    \label{fig:ConceptExtraction}
\end{figure*}
To validate our method, we assess its fidelity using the C-Deletion and C-Insertion metrics, as proposed by \citet{fel2023holistic}.
These methods provide a robust framework for evaluating the alignment between our explanation model and the original model's behavior by systematically modifying concept activations and observing the resulting impact on model predictions.

In the C-Deletion and C-Insertion metrics, concept vectors are removed or inserted in the order of their importance, and the Area Under the Curve (AUC) of the accuracy drop graph is measured. 
The importance score of a concept is calculated with Eq.~(\ref{eq:GIG_class}), as it is the most reliable CAT method \citep{fel2023holistic}.
For C-Deletion, a lower AUC indicates a more effective extraction method, as it signifies a greater impact on model performance when key concepts are removed.
Conversely, in C-Insertion, a higher AUC is preferable, reflecting a more accurate prediction when important concepts are introduced.
Finally, we measure the AUC difference to see the overall trends in every layer.


\textbf{Results \ } As shown in the top part of Fig.~\ref{fig:ConceptExtraction}, ours with Bisecting Clustering demonstrated consistently strong performance across most layers in both C-Deletion and C-Insertion.
Considering the fidelity metric, the difference between AUC(Insertion) and AUC(Deletion), ours outperforms the other methods, demonstrating its effectiveness in capturing and utilizing the most essential features of the model.
We observed that while SparseAutoEncoder (SAE) exhibited lowest fidelity in the earlier layers, it demonstrated exceptional performance in C-Insertion, achieving the highest AUC values in layer 4.
However, as seen in the bottom part of Fig.~\ref{fig:ConceptExtraction}, the concepts from SAE seem ambiguous compared to ours.


\subsubsection{Validation of Inter-layer Concept Attribution}
\label{val: attribution}
\begin{figure*}[t]
    \centering
    \includegraphics[width=\linewidth]{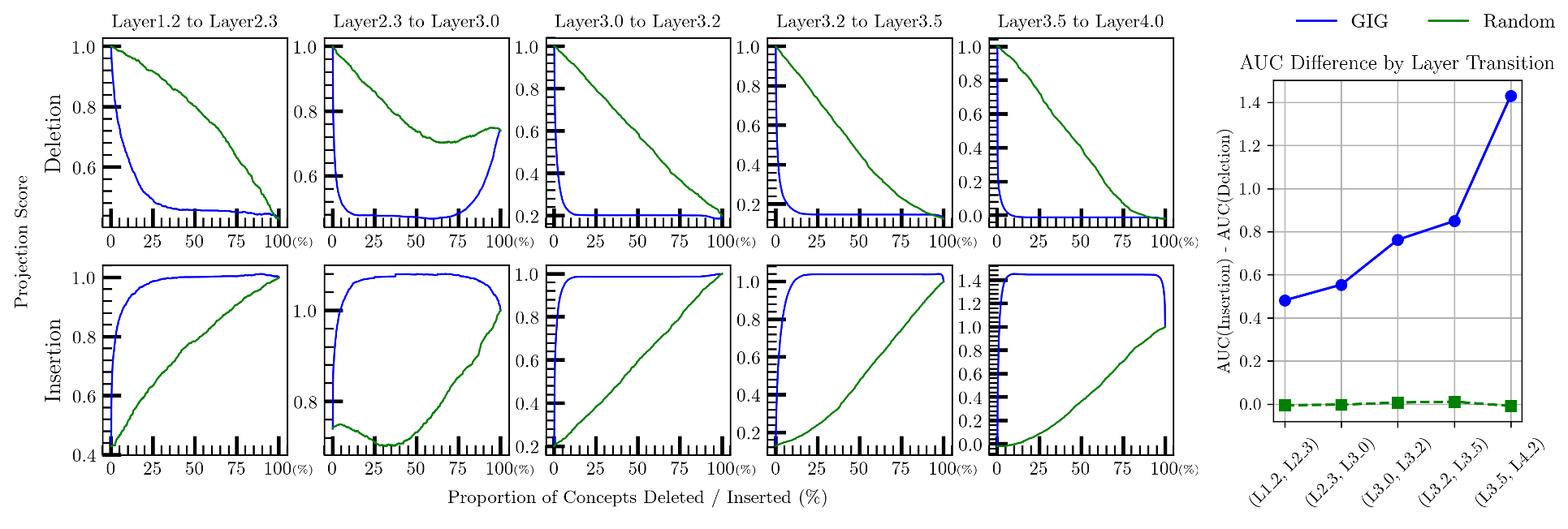}
    \caption{Left: Deletion (Top row) and Insertion (Bottom row) scores across consecutive layers in ResNet50. The blue curves represent our method (GIG), while the green curves denote random attribution. Our method consistently outperforms random attribution, as indicated by the significantly steeper decline in deletion scores and the sharper rise in insertion scores. Right: AUC Difference by Layer Transition. It quantifies our superiority, showing our AUC difference achieving substantially higher AUC differences across all layer transitions.}    
    \label{fig:Interlayer}
\end{figure*}
To validate the effectiveness of our concept attribution method, Generalized Integrated Gradient (GIG), we adapted the concept insertion and deletion strategies typically used in evaluating Concept ATtribution (CAT) methods.
The original C-Insertion and C-Deletion metrics quantify the relationship between the identified concept vectors and the target class. 
By systematically inserting or deleting concept vectors according to their attribution scores and observing changes in the target class score, we assess the validity of the concept vectors of CAT methods.

We extended this metric to validate the relationship between concept vectors in different layers.
As derived in Sec.~\ref{subsec: concept attribution}, the class label can be seen as the one-hot concept vector of the last layer after the fully-connected layer.
Therefore, we validated the efficacy of our inter-layer concept attribution method, observing the changes in the direction of the target concept vector in the subsequent layer by inserting or deleting concept vectors from a preceding layer.
Specifically, we deleted or inserted concept vectors from the source layer one by one and observed the changes in the output of the target layer (with dimensions $H^{b} \times W^{b} \times C$).
For instance, if we delete concept vectors related to the target concept ``dog nose" from the source layer in order of their GIG attribution, the output in the target layer corresponding to the ``dog nose" direction should decrease accordingly.
However, given that the actual region of ``dog nose" in the image may constitute only a small portion (e.g, less than 10\%) of the total image, removing the most relevant concept from the source layer will likely affect only 1-2 PFVs in the target layer.
Therefore, by deleting or inserting concepts in the source layer that most strongly contribute to the ``dog nose," and observing the change in the projection magnitude of the one PFV in the target layer that has the largest projection onto the ``dog nose" direction, we can determine whether the attribution computed by GIG is valid.

To this end, we plotted the curve of the normalized maximum projection values of the PFVs in the target layer onto the target concept vector direction.
Specifically, during the Insertion/Deletion processes, the maximum projection values at each step were normalized by the original maximum projection value prior to any Insertion or Deletion.
We refer to this normalized value as the projection score, and this metric as Interlayer Insertion/Deletion. 

We conducted the Inter-layer Deletion/Insertion experiments on both GIG attribution and random attribution.
For random attribution, we deleted or inserted concept vectors in a random order.

\textbf{Results. \ }
The left plot in Fig.~\ref{fig:Interlayer} displays the Inter-layer Deletion/Insertion curves between various blocks, specifically [Layer1.2, Layer2.3, Layer3.0, Layer3.2, Layer3.5, Layer4.2].
This experiment was conducted on the average projection score of the five most important target layer concepts across 20 random images from the ImageNet validation set.
As expected, the curve for GIG attribution shows a rapid decrease/increase during deletion/insertion, outperforming the random attribution.

Interestingly, Insertion curves of GIG sometimes exceed 1, indicating that the insertion of only the positively attributed concepts leads to a higher maximum projection value in the target layer output than that in the original output. 
The projection score returns to 1, as the original layer output is restored after the negatively attributed concepts are inserted.

The right plot in Fig.~\ref{fig:Interlayer} shows the difference in AUC between GIG and random attribution. 
The significant AUC difference in GIG validates our method, demonstrating its effectiveness in accurately attributing the relationship between concept vectors across whole layers.


\section{Conclusion}
In this paper, we firstly present a novel approach for extracting and attributing concepts within image models, enhancing interpretability through a comprehensive layer-wise analysis.
Unlike existing methods that often confine their explanations to specific classes, our approach provides a comprehensive understanding by analyzing shared concepts throughout the dataset.
The shift from class-specific to dataset-wide explanations represents a significant advancement in the field of XAI in image models, allowing for a more holistic understanding of model behavior.

With the dataset of PFV and ERF, we propose a pipeline that systematically decompose PFVs into meaningful concept vectors, and further attribute these concepts across layers using the Generalized Integrated Gradient (GIG) method.
With our method, we can reveal how concepts evolve and influence decisions across different layers of the network.
Through extensive qualitative and quantitative analyses, we demonstrate the effectiveness of our method in both accurately capturing and utilizing essential features.

Given its potential for broad applicability, we can extend our method to other deep learning architectures, such as Transformer models.
Additionally, the implications of analyzing entire datasets rather than focusing solely on class-specific explanations could be more thoroughly investigated. 
We believe that our approach opens a new avenues for interpretability in image models, broadening the perspective of XAI.

\subsubsection*{Author Contributions}
If you'd like to, you may include  a section for author contributions as is done
in many journals. This is optional and at the discretion of the authors.

\subsubsection*{Acknowledgments}
Use unnumbered third level headings for the acknowledgments. All
acknowledgments, including those to funding agencies, go at the end of the paper.

\bibliography{iclr2024_conference}
\bibliographystyle{iclr2024_conference}

\end{document}